\documentclass[10pt,twocolumn,letterpaper]{article}

\usepackage{iccv}

\usepackage{times}
\usepackage{epsfig}
\usepackage{graphicx}
\usepackage{amsmath}
\usepackage{amssymb}
\usepackage{booktabs}
\usepackage[misc]{ifsym}

\usepackage[pagebackref=true,breaklinks=true,letterpaper=true,colorlinks,bookmarks=false]{hyperref}

\usepackage[capitalize]{cleveref}
\crefname{section}{Sec.}{Secs.}
\Crefname{section}{Section}{Sections}
\Crefname{table}{Table}{Tables}
\crefname{table}{Tab.}{Tabs.}

\usepackage{caption}
\usepackage{subcaption}
\usepackage{pifont} 
\usepackage{multirow, overpic, textpos, array}
\usepackage{makecell}

\usepackage{color, colortbl}
\definecolor{Gray}{gray}{0.9}
\newcommand{\GZTUPDATED}{}

\newlength\savewidth\newcommand\shline{\noalign{\global\savewidth\arrayrulewidth
  \global\arrayrulewidth 1pt}\hline\noalign{\global\arrayrulewidth\savewidth}}
\newcommand{\tablestyle}[2]{\setlength{\tabcolsep}{#1}\renewcommand{\arraystretch}{#2}\centering\footnotesize}
\renewcommand{\paragraph}[1]{\vspace{1.25mm}\noindent\textbf{#1}}
\newcommand\blfootnote[1]{\begingroup\renewcommand\thefootnote{}\footnote{#1}\addtocounter{footnote}{-1}\endgroup}

\newcommand{\x}{$\times$}

\newcolumntype{x}[1]{>{\centering\arraybackslash}p{#1pt}}
\newcolumntype{a}[1]{>{\columncolor{Gray}\centering\arraybackslash}p{#1pt}}
\newcolumntype{y}[1]{>{\raggedright\arraybackslash}p{#1pt}}
\newcolumntype{z}[1]{>{\raggedleft\arraybackslash}p{#1pt}}

\newcommand{\ImageNetOneKClassification}{
\begin{table}[b]
    \centering
    \footnotesize
    \setlength{\tabcolsep}{2.0pt}
    \renewcommand\arraystretch{1.05}
    \vspace{-1.em}
    \begin{tabular}{c|c|ccc}
        method & top-1 & FLOPs & \#params & throughput (img/s)\\
        \shline
        ResNet-50~\cite{wightman2021resnet}
        & 80.4 & 4.1G & 26M & 1179 \\
        ResNet-101~\cite{DBLP:conf/cvpr/HeZRS16}
        & 81.5 & 7.9G & 45M & 691 \\
        RegNetY-8G~\cite{DBLP:conf/cvpr/RadosavovicKGHD20}
        & 81.7 & 8.0G & 39M & 592 \\
        EfficientNet-B3$^*$~\cite{effnet}
        & 81.6 & 1.8G & 12M & 745\\
        \hline
        DeiT-S~\cite{deit}
        & 79.8 & 4.6G & 22M & 983 \\
        DeiT-B~\cite{deit}
        & 81.8 & 17.5G & 86M & 306 \\
        Swin-T~\cite{swin}
        & 81.3 & 4.5G & 29M & 726 \\
        Swin-S~\cite{swin}
        & 83.0 & 8.7G & 50M & 437 \\
        Perceiver~\cite{perceiver}
        & 78.0 & 707G & 45M & 17\\
        Perceiver IO~\cite{perceiverio}
        & 82.1 & 369G & 49M & 30\\
        \hline
        SparseFormer-T
        & 81.0 & 2.0G & 32M & 1270 \\
        SparseFormer-S
        & 82.0 & 3.8G & 48M & 898  \\
        SparseFormer-B
        & 82.6 & 7.8G & 81M & 520 \\
        \end{tabular}
    \vspace{-0.0em}
    \caption{
    Comparison of different networks on ImageNet-1K classification. The input resolution is $224^2$ except for  Efficient-B3 with the resolution $300^2$.
    The throughput is measured on a single V100 GPU, following~\cite{swin}
    }
    \label{table:imagenetclassification}
\end{table}
}

\newcommand{\ScalingUp}{
\begin{table}[h]
    \centering
    \footnotesize
    \setlength{\tabcolsep}{2.0pt}
    \renewcommand\arraystretch{1.05}
    \vspace{-1em}
    \begin{tabular}{cccccc}
        variant & pre-train & resolution & top-1 & FLOPs & throughput (img/s)\\
        \shline
        B & 1K & $224^2$ & 82.6 & 7.8G & 520 \\
        \hline
        B & 21K & $224^2$ & 83.6 & 7.8G & 520 \\
        B & 21K & $384^2$ & 84.1 & 8.2G & 444 \\
        B & 21K & $512^2$ & 84.0 & 8.6G & 419 \\
        \hline
        B, $N=144\uparrow$ & 21K & $384^2$ & 84.6 & 14.2G & 292 \\
        B, $N=196\uparrow$ & 21K & $384^2$ & 84.8 & 19.4G & 221 
        \end{tabular}
    \vspace{-0.0em}
    \caption{
    \textbf{Scaling up} of SparseFormer-B. 
    Except for the 1K entry, all follow first the same pre-training on ImageNet-21K ($224^2$ input, $81$ tokens) and then individual fine-tuning on ImageNet-1K.
    \vspace{-1.5em}
    }
    \label{table:twentyonekandscalingup}
\end{table}
}

\newcommand{\VideoUnderstanding}{
\begin{table}[b]
    \centering
    \footnotesize
    \setlength{\tabcolsep}{1.5pt}
    \renewcommand\arraystretch{1.05}
    \vspace{-1em}
    \begin{tabular}{c|ccccc}
        method & top-1  & pre-train & \#frames & GFLOPs & \#params \\
        \shline
        NL I3D~\cite{nonlocal} & 77.3  & ImageNet-1K & 128 & 359\x10\x3  & 62M \\
        SlowFast~\cite{slowfast} & 77.9 & - & 8+32 & 106\x10\x3 & 54M \\
        TimeSFormer~\cite{timesformer}   &  75.8 & ImageNet-1K & 8 & 196\x1\x3 &  121M \\
        Video Swin-T~\cite{liu2021video} & 78.8  & ImageNet-1K & 32 & 88\x4\x3 & 28M \\
        ViViT-B FE~\cite{vivit}  & 78.8   &  ImageNet-21K & 32 & 284\x4\x3 & 115M \\
        MViT-S~\cite{mvit}  & 76.0  & - & 16 & 33\x5\x1 & 26M \\
        MViT-B~\cite{mvit}  & 78.4  & - & 16 & 71\x5\x1 & 37M \\
        \hline
        VideoSF-T
        & 77.9  & ImageNet-1K & 32 & \GZTUPDATED 22\x4\x3  & 31M \\
        VideoSF-S
        & 79.1 & ImageNet-1K & 32 & \GZTUPDATED 38\x4\x3 & 48M \\
        VideoSF-B
        & 79.8    &  ImageNet-21K & 32 & \GZTUPDATED 74\x4\x3  & 81M \\
        \end{tabular}
        \vspace{-0.0em}
        \caption{\textbf{Comparison with well-established video classification methods} on Kinetics-400. The GFLOPs is in the format of a single view \x\ the number of views. ``N/A'' indicates the numbers are not available.
        \vspace{-0.0em}
        }
        \label{table:video}
\end{table}
}

\newcommand{\VideoUnderstandingAblationOnInflation}{
\begin{table}[h]
    \centering
    \footnotesize
    \setlength{\tabcolsep}{4.0pt}
    \renewcommand\arraystretch{1.05}
    \vspace{-1em}
    \begin{tabular}{c|cc}
        inflation & top-1  & GFLOPs \\
        \shline
        1 & 69.5  & 7\x 4\x 3 \\
        4 & 74.7  & 13\x 4\x 3 \\
        \rowcolor{Gray} 8 & 77.9 & 22\x 4\x 3 \\
        16 & 78.2 & 32\x 4\x 3
    \end{tabular}
    \vspace{-0.0em}
    \caption{\textbf{Different inflation rates} of VideoSparseFormer-T on Kinetics-400.
    \vspace{-1em}
    }
    \label{table:videoInflation}
\end{table}
}

\newcommand{\ModelConfig}{
\begin{table}[h]
    \centering
    \footnotesize
    \setlength{\tabcolsep}{2.0pt}
    \renewcommand\arraystretch{1.0}
    \vspace{-0.5em}
    \begin{tabular}{l|cccccc}
        variant &
        \makecell[c]{\#tokens \\ $N$} &
        \makecell[c]{foc. dim\\ $d_{f}$} &
        \makecell[c]{cort. dim\\ $d_{c}$} &
        \makecell[c]{cort. stage\\ $L_{c}$} &
        FLOPs &
        \#params \\
        \shline
        tiny (T) & 49 & 256 & 512 & 8  & 2.0G & 32M\\
        small (S) & 64 & 320 & 640 & 8  & 3.8G & 48M\\
        base (B) & 81 & 384 & 768 & 10 & 7.8G & 81M\\
        \end{tabular}
    \vspace{-0em}
    \caption{\textbf{Configurations} of SparseFormer variants. FLOPs is calculated with the input image size $224^2$.
    }
    \vspace{-0em}
    \label{table:modelConfig}
\end{table}
}

\newcommand{\Ablation}{
    \begin{table*}[t]
        \centering
        \subfloat[\footnotesize
        \textbf{The number of latent tokens} $N$
        \label{tab:tokennumber}
        ]{
        \centering
        \begin{minipage}{\linewidth}{\begin{center}
        \tablestyle{1pt}{1.1}
        \footnotesize
        \begin{tabular}{x{32}|x{32}x{32}x{32}x{32}a{32}x{32}x{32}}
        $N$ & 9 & 16 & 25 & 36 & 49 & 64 & 81 \\
        \shline
        top-1 & 74.5 & 77.4 & 79.3 & 80.1 & 81.0 & 81.4 & 81.9 \\
        GFLOPs & 0.5 & 0.8 & 1.1 & 1.6 & 2.0 & 2.7  &3.3 \\
        \end{tabular}
        \end{center}}\end{minipage}
        }\vspace{0.2em}
        \\
        
        \subfloat[\footnotesize
        \textbf{Repeats of the focusing Transformer}, $L_s$
        \label{tab:focusingstage}
        ]{
        \centering
        \begin{minipage}{0.23\linewidth}{\begin{center}
        \tablestyle{1pt}{1.1}
        \footnotesize
        \begin{tabular}{x{32}x{32}x{32}x{32}}
        $L_s$ & top-1 & GFLOPs \\
        \shline
        nil & 77.8 & 1.6 \\
        1   & 79.7 & 1.7 \\
        \rowcolor{Gray}
        4   & 81.0 & 2.0 \\ 
        8   & 81.0 & 2.5 \\ 
        \end{tabular}
        \end{center}}\end{minipage}
        }
        \hfill
        \subfloat[\footnotesize
        \textbf{The number sampling points} $P$ for a token in a sampling stage
        \label{tab:samplingpoints}
        ]{
        \centering
        \begin{minipage}{0.23\linewidth}{\begin{center}
        \tablestyle{1pt}{1.1}
        \footnotesize
        \begin{tabular}{x{32}x{32}x{32}x{32}}
        $P$ & top-1 & GFLOPs \\
        \shline
        16 & 80.3 & 1.9 \\ 
        \rowcolor{Gray}
        36 & 81.0 & 2.0 \\ 
        64 & 81.3 & 2.3 \\ 
          \\
        \end{tabular}
        \end{center}}\end{minipage}
        }%
        \hfill
        \subfloat[\footnotesize
        \textbf{Image features} to be sampled
        \label{tab:imagefeatures}
        ]{
        \centering
        \begin{minipage}{0.23\linewidth}{\begin{center}
        \tablestyle{1pt}{1.1}
        \footnotesize
        \begin{tabular}{x{50}x{32}x{32}x{32}x{32}a{32}x{32}x{32}}
        img feat. & top-1 & GFLOPs \\ 
        \shline
        RGB          & fail & 1.5 \\ 
        ViT/8-embed  & 78.4 & 1.9 \\ 
        \rowcolor{Gray}
        early conv & 81.0 & 2.0 \\ 
        \\
        \end{tabular}
        \end{center}}\end{minipage}
        }%
        \hfill
        \subfloat[\footnotesize
        \textbf{How to decode} sampled features
        \label{tab:howtodecode}
        ]{
        \centering
        \begin{minipage}{0.23\linewidth}{\begin{center}
        \tablestyle{1pt}{1.1}
        \footnotesize
        \begin{tabular}{x{50}x{32}x{32}x{32}x{32}a{32}x{32}x{32}}
        decode & top-1 & GFLOPs \\ 
        \shline
        linear        & 78.5 & 1.9 \\ 
        static, mix   & 80.1 & 1.9 \\ 
        \rowcolor{Gray}
        adaptive, mix & 81.0 & 2.0 \\ 
        \\
        \end{tabular}
        \end{center}}\end{minipage}
        }%

        \vspace{-0.4em}
        \caption{\textbf{Ablations study} on SparseFormer.
        The default choice for SparseFormer is colored \colorbox{Gray}{gray}.
        \vspace{-1em}
        }
        \label{tab:ablations}
        \end{table*}
}

\newcommand{\BoxInit}{
\begin{table}[h]
    \centering
    \small
    \setlength{\tabcolsep}{12.0pt}
    \renewcommand\arraystretch{1.0}
    \begin{tabular}{c|c}
        width and height initialization & top-1 \\
        \shline
        {half, $0.5\times 0.5$} & 81.0 \\
        {cell, $1/\sqrt{N}\times 1/\sqrt{N}$} & 81.0 \\
        {whole, $1.0\times 1.0$} & 80.2 \\
        \end{tabular}
    \vspace{-0em}
    \caption{Alternative ways to initialize the token height and width for SparseFormer-T. $N$ is the number of latent tokens for the `cell' initialization. The `cell' initialization tiles RoIs without overlapping over the image. The `whole' initialization is with all token RoIs centered at the image center.
    }
    \vspace{-0em}
    \label{tab:box_init}
\end{table}
}

\newcommand{\Det}{
\begin{table}[h]
    \centering
    \small
    \setlength{\tabcolsep}{2.0pt}
    \renewcommand\arraystretch{1.1}
    \begin{tabular}{c|c|cccccc}
        detector & GFLOPs & AP & AP$_{50}$ & AP$_{75}$ & AP$_s$ & AP$_m$ & AP$_l$  \\
        \shline
        DETR & 86 & 42.0 & 62.4 & 44.2 & 20.5 & 45.8 & 61.1 \\
        DETR-DC5 & 187 & 43.3 & 63.1 & 45.9 & 22.5 & 47.3 & 61.1 \\
        \shline
        SparseFormer-S & 27 & 26.4 & 43.8 & 26.6 & 8.3 & 26.0 & 45.0 \\
        \end{tabular}
    \vspace{-0em}
    \caption{Detection performance of SparseFormer-S on MS COCO~\cite{DBLP:conf/eccv/LinMBHPRDZ14} \texttt{val} set.
    }
    \vspace{-0em}
    \label{tab:det}
\end{table}
}

\newcommand{\Seg}{
\begin{table}[h]
    \centering
    \small
    \setlength{\tabcolsep}{2.0pt}
    \renewcommand\arraystretch{1.1}
    \begin{tabular}{c|c|cccccc}
        segmentor & GFLOPs & mIoU & mAcc  \\
        \shline
        Swin-T~\cite{swin} + UperNet~\cite{UperNet} & 236 & 44.4 & 56.0 \\
        \shline
        SF-T w/ 49 tokens & 33 & 36.1 & 46.0 \\
        SF-T w/ 256 tokens & 39 & 42.9 & 53.7 \\
        SF-T w/ 400 tokens & 43 & 43.5 & 54.7 \\
        \end{tabular}
    \vspace{-0em}
    \caption{Semantic segmentation performance of SparseFormer-T on Ade20K~\cite{ade20k} validation set. The GFLOPs are computed with $512\times 512$ input resolution.
    }
    \vspace{-0em}
    \label{tab:seg}
\end{table}
}
\usepackage{caption}
\usepackage{subcaption}

\newcommand{\FigureIntro}{
    \begin{figure}[t]
        \centering
        \includegraphics[width=\columnwidth]{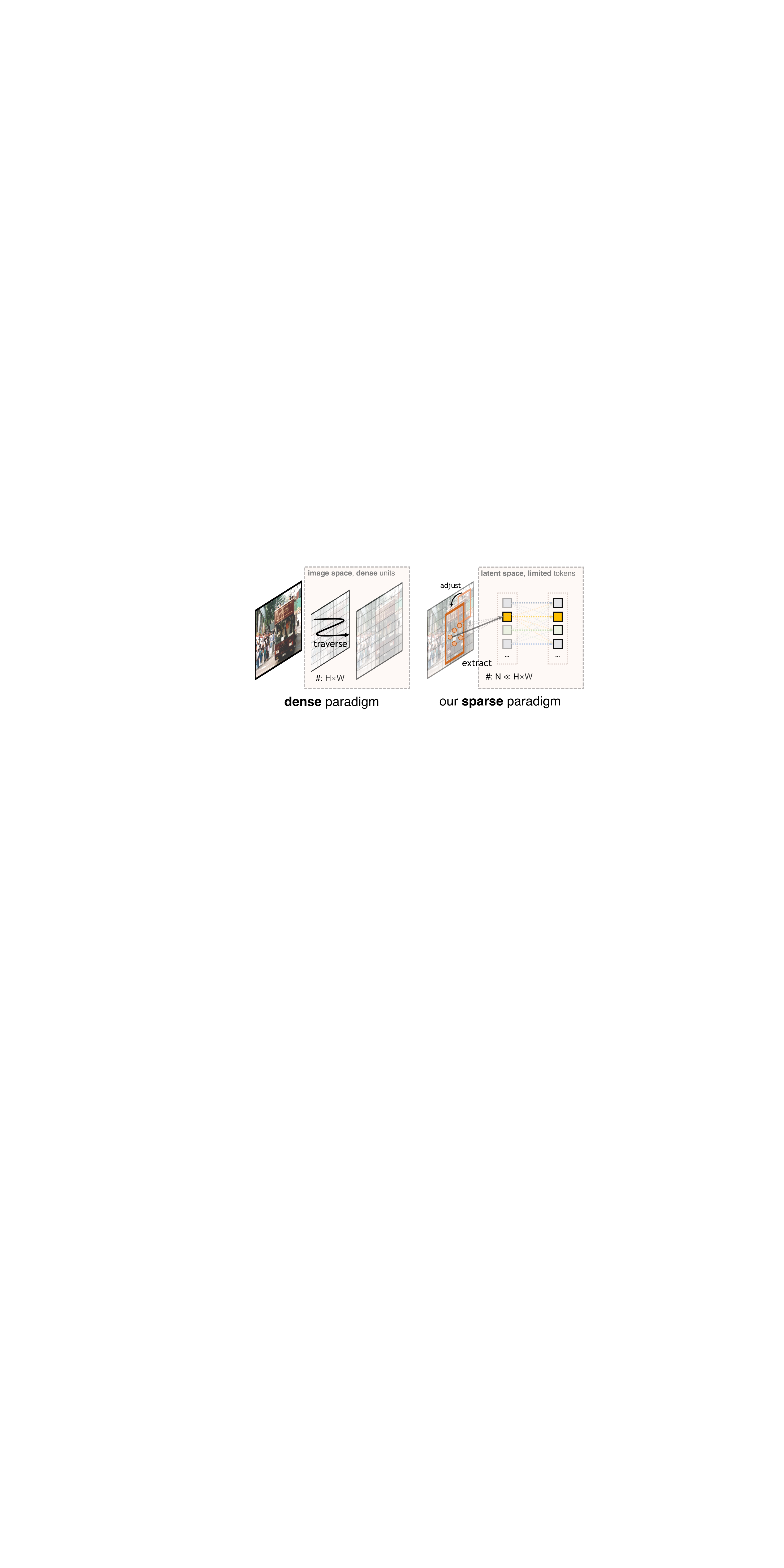}
        \caption{\textbf{Dense} versus our proposed \textbf{sparse} paradigm for visual recognition.
        The dense paradigm requires traversing $H\times W$ units to perform convolution or attention, while our proposed sparse paradigm performs transformers over only $N$ latent tokens where $N\ll H\times W$. \vspace{-1em} }
        \label{fig:intro}
    \end{figure}
        
}

\newcommand{\FigureArch}{
    \begin{figure*}[ht]
        \centering
        \includegraphics[width=0.9\textwidth]{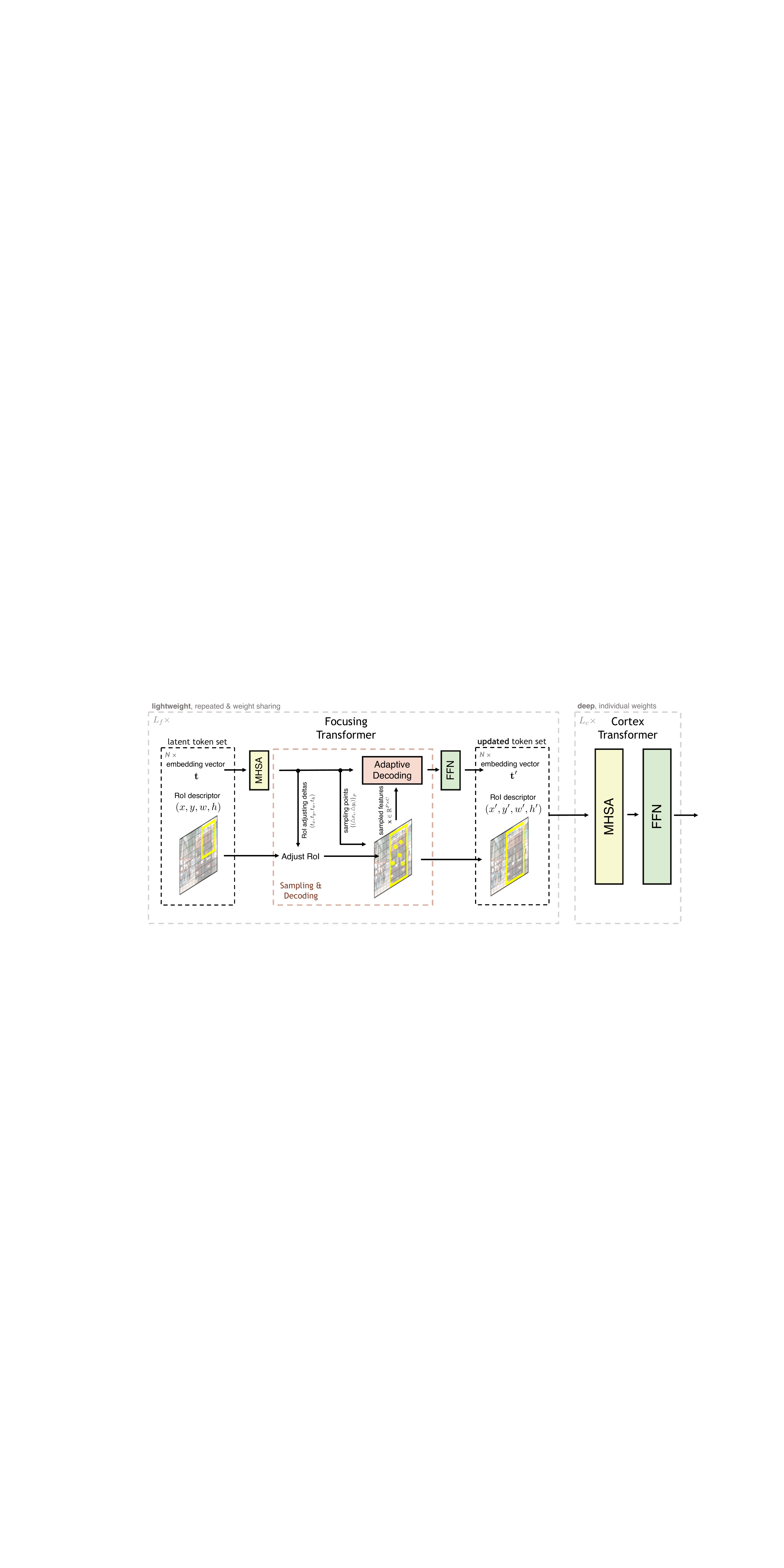}
        \caption{\textbf{The overall architecture} of the proposed SparseFormer.
        The MHSAs and FFNs are multi-head self-attention over latent tokens and feed-forward networks applied on these tokens, respectively. 
        Note that different from the illustration aimed for details, the cortex Transformer is actually much deeper and wider than the focusing Transformer. 
        We omit layer normalization here for more clarity.
        All of these operations are performed in the latent token space, except for image feature sampling in the original image space.\vspace{-0.5em}
        }
        \label{fig:arch}
    \end{figure*}
        
}

\newcommand{\FigureVis}{
    \begin{figure}[t]
        \centering
        \includegraphics[width=0.9\columnwidth]{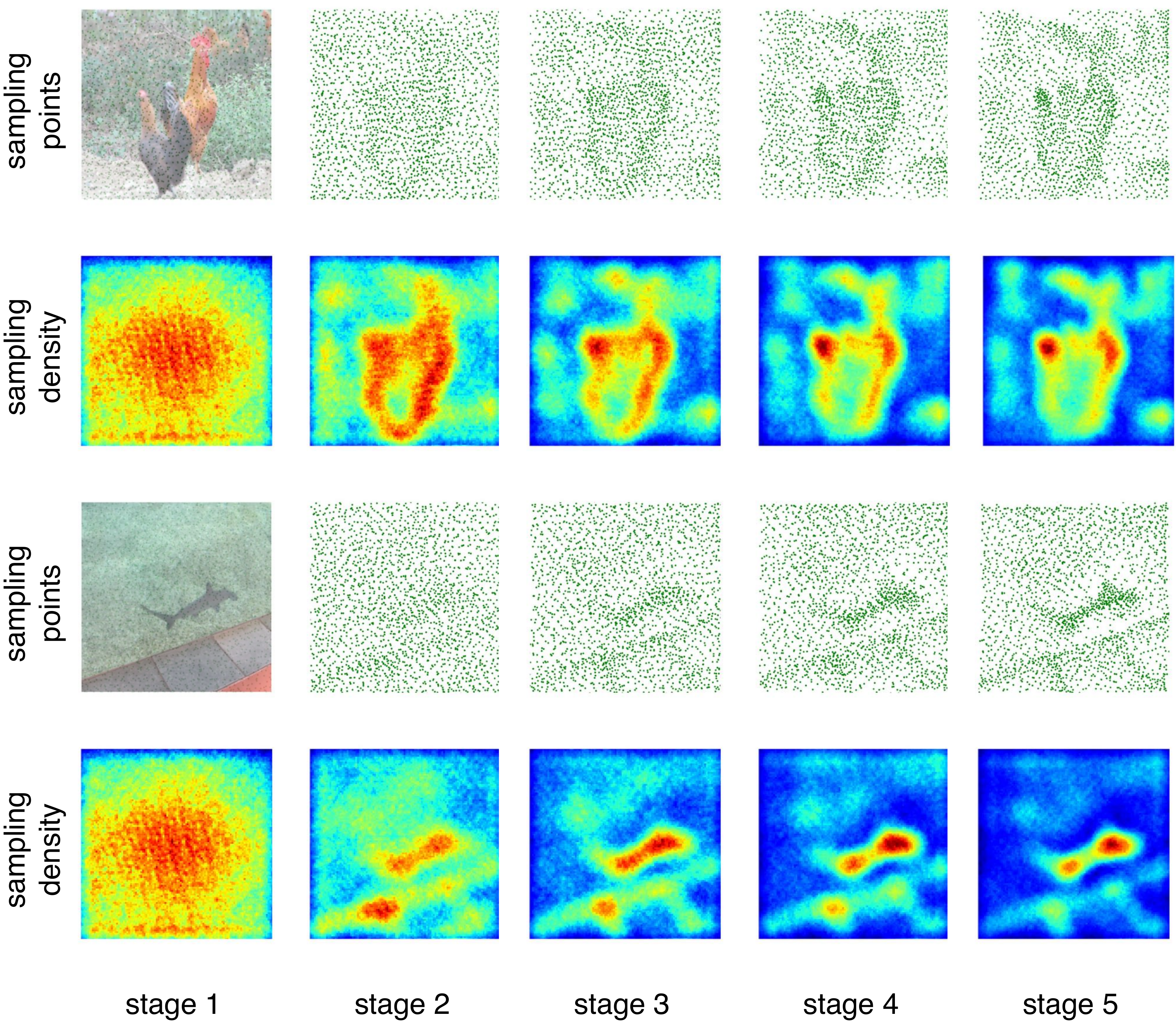}
        \caption{\textbf{Visualizations} of sampling points and their sampling density maps across sampling stages in SparseFormer-S. Stage 1-4 refer to the feature sampling in the focusing Transformer, and Stage 5 refers to the cortex Transformer. Better view with zoom-in.\vspace{-1em}}
        \label{fig:vis}
    \end{figure}
        
}

\newcommand{\FigureSuppOne}{
    \begin{figure*}[t]
        \centering
        \includegraphics[width=0.9\textwidth]{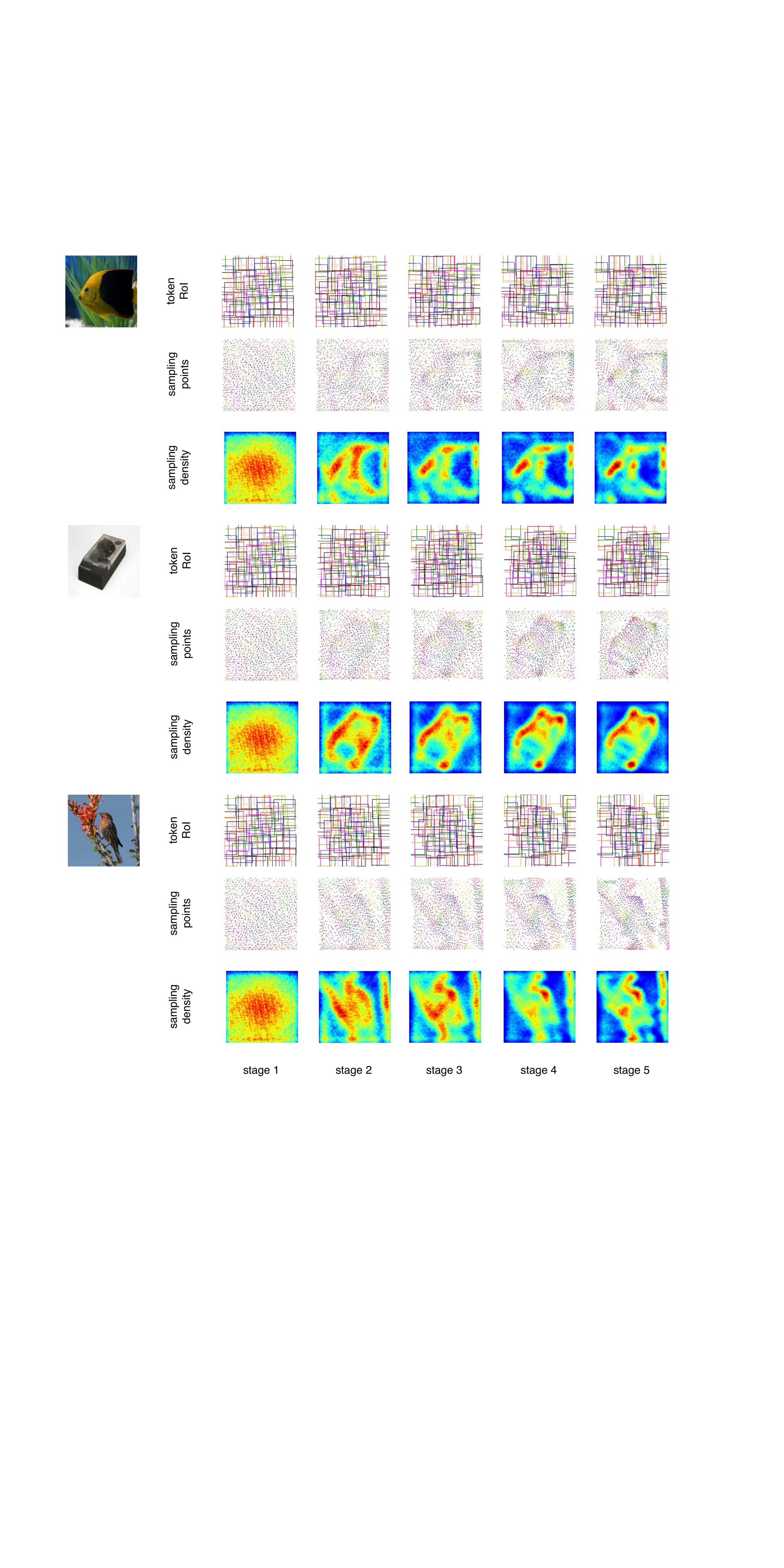}
        \caption{
            \textbf{More visualizations} (cont'd).
        }\label{fig:two}
    \end{figure*}
}

\newcommand{\FigureSuppToken}{
    \begin{figure*}[t]
        \centering
        \includegraphics[width=0.9\textwidth]{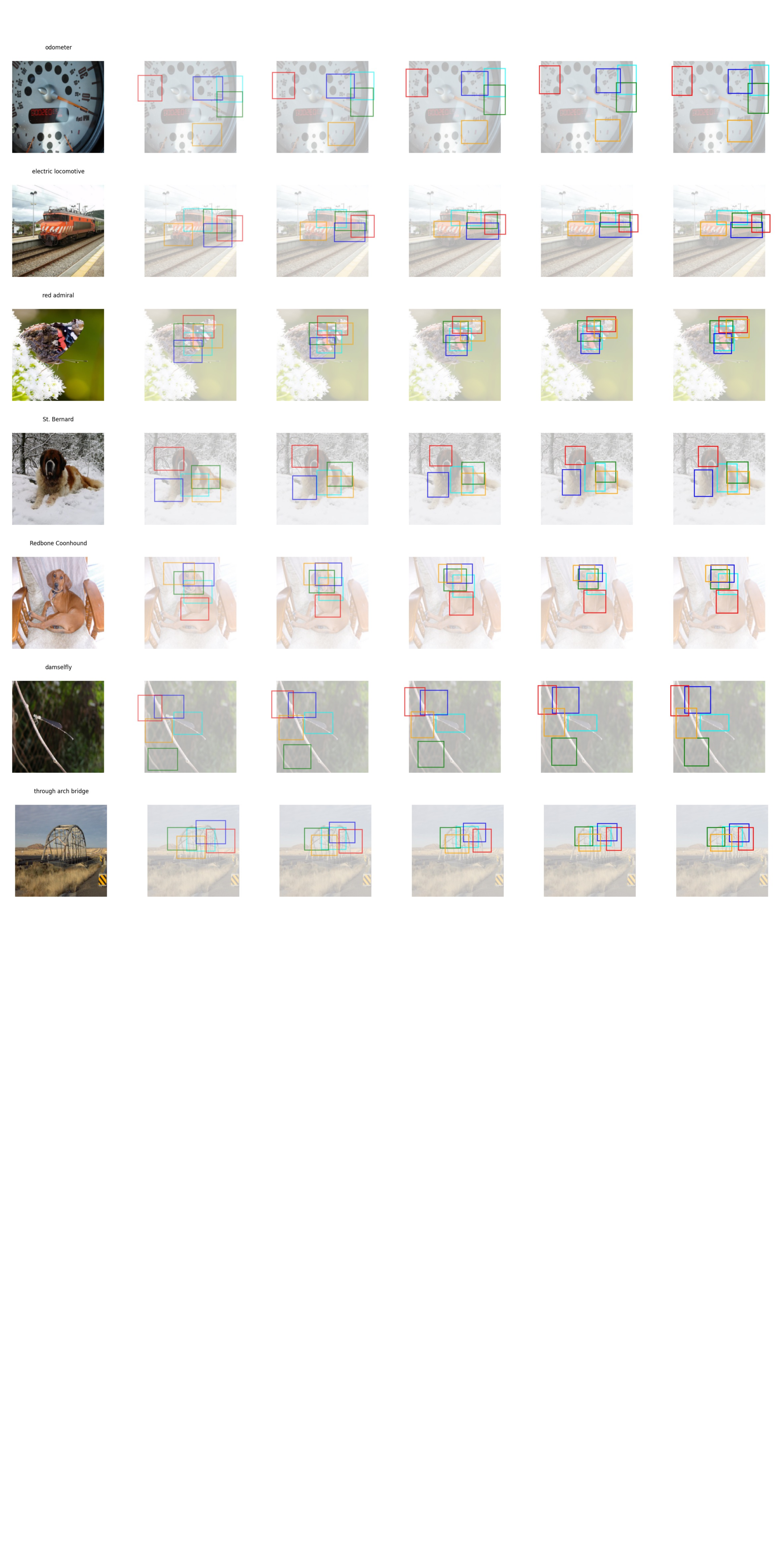}
        \caption{
            Visualizations on top-5 tokens responding to the ground-truth category across stages.
        }
        \label{fig:token}
    \end{figure*}
}

\newcommand{\FigureSuppDis}{
    \begin{figure*}[t]
        \centering
        \includegraphics[width=0.9\textwidth]{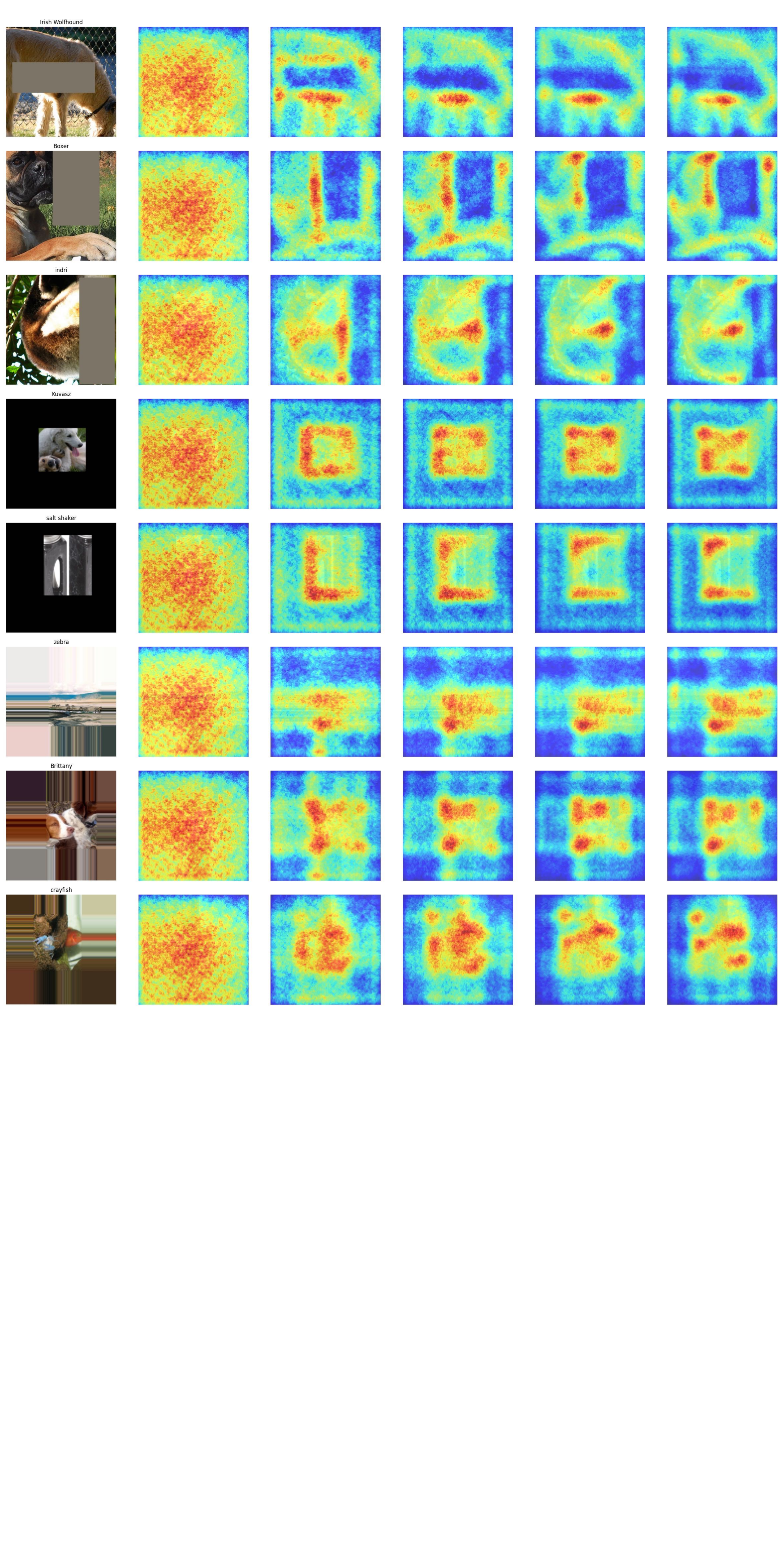}
        \caption{
            Visualizations on sampling density maps when disturbing input images.
        }
        \label{fig:distu}
    \end{figure*}
}

\iccvfinalcopy 

\def\iccvPaperID{****} 

\ificcvfinal\fi

\begin{document}

\title{SparseFormer: Sparse Visual Recognition via Limited Latent Tokens}

\author{
Ziteng Gao\textsuperscript{1} \quad \quad
Zhan Tong\textsuperscript{2} \quad \quad
Limin Wang\textsuperscript{3} \quad \quad
Mike Zheng Shou\textsuperscript{1 \Letter}
\vspace{0.2em}
\\
\textsuperscript{1}Show Lab, National University of Singapore \hfill \hspace{0.2em}
\textsuperscript{2}Tencent AI Lab \hfill \hspace{0.2em}
\textsuperscript{3}Nanjing University
}
\maketitle

\begin{abstract}
Human visual recognition is a \emph{sparse} process, where only a few salient visual cues are attended to rather than traversing every detail uniformly. 
However, most current vision networks follow a \emph{dense} paradigm, processing every single visual unit (\eg, pixel or patch) in a uniform manner.
In this paper, we challenge this dense paradigm and present a new method, coined \emph{{SparseFormer}}, to imitate human's \emph{sparse} visual recognition in an end-to-end manner.
SparseFormer learns to represent images using a highly limited number of tokens (down to $49$) in the latent space with sparse feature sampling procedure instead of processing dense units in the original pixel space. 
Therefore, SparseFormer circumvents most of dense operations on the image space and has much lower computational costs.
Experiments on the ImageNet classification benchmark dataset show that SparseFormer achieves performance on par with canonical or well-established models while offering better accuracy-throughput tradeoff.
Moreover, the design of our network can be easily extended to the video classification with promising performance at lower computational costs.
We hope that our work can provide an alternative way for visual modeling and inspire further research on sparse neural architectures. The code will be publicly available at \url{https://github.com/showlab/sparseformer}.

\end{abstract}
\blfootnote{ \Letter: Corresponding Author.}

\section{Introduction}
Designing neural architectures for visual recognition has long been an appealing yet challenging topic in the computer vision community.
Convolutional neural networks (CNNs)~\cite{DBLP:conf/nips/KrizhevskySH12,DBLP:conf/cvpr/HeZRS16,DBLP:journals/corr/HuangLW16a,DBLP:conf/cvpr/RadosavovicKGHD20,convnext} use convolutional filters on every unit of images or feature maps to build features.
Recently proposed vision transformers~\cite{vit,deit,swin,pvt} use attention operation on each patch or unit to dynamically interact with other units to mimic human attention mechanism.
Both convolution-based and Transformer-based architectures need to traverse every unit, like pixel or patch on the grid map, to densely perform operations.
This dense per-unit traversal originates from sliding windows~\cite{lbp}, reflecting the assumption that foreground objects may appear uniformly with respect to spatial locations in an image.

\FigureIntro

However, as humans, we do not need to examine every detail in a scene to recognize it.
Instead, we are fast enough to first roughly find discriminative regions of interest with several glimpses and then recognize textures, edges, and high-level semantics within these regions~\cite{doi:10.1146/annurev.ne.18.030195.001205,10.7551/mitpress/9780262514620.001.0001,Itti2001,Rensink2000TheDR}.
This contrasts greatly with existing visual networks, where the convention is to exhaustively traverse every visual unit. 
The dense paradigm incurs soaring computational costs with larger input resolutions and does not directly provide details about what a vision model is looking at in an image.

In this paper, we propose a new vision architecture, coined \textbf{SparseFormer}, to explore \emph{sparse visual recognition} by explicitly imitating the perception mechanism of human eyes.
Specifically, SparseFormer learns to represent an image by latent transformers along with a highly limited number of tokens (\eg, down to $49$) in the latent space \emph{from the very beginning}.
Each latent token is associated with a region of interest (RoI) descriptor, and the token RoI can be refined across stages.
Given an image, SparseFormer first extracts image features by a lightweight early convolution module. 
A latent focusing transformer adjusts token RoIs to focus on foregrounds and sparsely extracts image features according to these token RoIs to build latent token embeddings in an iterative manner.
SparseFormer then feeds tokens with these region features into a larger and deeper network, \ie a standard transformer encoder in the latent space, to enable precise recognition.
The transformer operations are only applied over the limited tokens in the latent space.
Since the number of latent tokens is \emph{highly limited} (\eg, $49$) and the feature sampling procedure is \emph{sparse} (\ie, based on direct bilinear interpolation), it is reasonable to call our architecture a \emph{sparse} approach for visual recognition.
The overall computational cost of SparseFormer is \emph{almost irrelevant} to the input resolution except for the early convolution part, which is lightweight in our design.
Moreover, SparseFormer can be supervised \emph{solely with classification signals} in an \emph{end-to-end} manner without requiring separate prior training with localizing signals.

As an initial step to sparse visual recognition, the aim of SparseFormer is to explore an alternative paradigm for vision modeling, rather than state-of-the-art results with bells and whistles.
Nonetheless, SparseFormer still delivers very promising results on the challenging ImageNet classification benchmark, on par with dense counterparts but at lower computational costs.
Since most operators of SparseFormer are performed on tokens in the latent space rather than the dense image space, and the number of tokens is limited, the memory footprints are lower and throughputs are higher compared to dense architectures, yielding a better accuracy-throughput trade-off, especially in the low-compute region.
For instance, with ImageNet 1K training, the tiny variant of SparseFormer with $2.0$G FLOPs gives $81.0$ top-1 accuracy at a throughput of $1270$ images/s, while Swin-T with $4.5$G FLOPs achieves $81.3$ at a lower throughput of $726$ images/s.
Visualizations of SparseFormer validate its ability to differentiate foregrounds from backgrounds in an end-to-end manner using only classification signals. 
Furthermore, we also investigate several scaling-up strategies of SparseFormer for better performance.

The simple design of SparseFormer can also be easily extended to video classification, which is more data-intensive and computationally expensive for dense vision models but is suitable for the SparseFormer architecture.
Experimental results on the challenging video classification Kinetics-400~\cite{kinetics} benchmark show that our extension of SparseFormer in video classification yields promising performance with lower computation than dense architectures.
This highlights the efficiency and effectiveness of the proposed sparse vision architecture with denser input data.

\section{Related Work}
\noindent\textbf{Convolutional neural networks.}
AlexNet~\cite{DBLP:conf/nips/KrizhevskySH12} pioneered convolutional neural networks for general visual recognition by using stacked local convolutions on dense pixels to build semantic features.
Since then, CNNs have dominated visual understanding with improvement on several aspects, \eg, pathway connections~\cite{DBLP:conf/cvpr/HeZRS16,DBLP:journals/corr/HuangLW16a}, convolutional filter configurations~\cite{DBLP:conf/cvpr/RadosavovicKGHD20,effnet,convnext}, multi branches~\cite{DBLP:conf/cvpr/SzegedyVISW16,DBLP:conf/cvpr/SzegedyLJSRAEVR15}, normalizations~\cite{DBLP:conf/icml/IoffeS15,DBLP:journals/ijcv/WuH20}.
To cover discriminative details, all convolutional networks require dense convolution applied over pixels. Although pooling layers are used to downsize feature maps spatially, exhaustive convolution over dense units still slows down networks, particularly for large inputs.
The case becomes more severe when dealing with data-intensive video inputs, as the dense paradigm for videos introduces a significantly heavier computational burden~\cite{c3d,kinetics,kinetics}. Even with adopted factorized convolution~\cite{R21D,p3d,s3d} or sampling strategies~\cite{tsn,slowfast}, the computational burden is still significant.

\noindent\textbf{(Efficient) vision transformers.}
Recently proposed Vision Transformer (ViT)~\cite{vit} makes use of a fully Transformer-based architecture originating in NLP~\cite{DBLP:conf/nips/VaswaniSPUJGKP17} to enable visual recognition with a larger model capacity.
ViT first splits an input image into patches with a moderate number (\eg, 196) and feeds these patches as tokens into a standard Transformer for recognition.
Attention operators inside ViT learn dynamic and long-term relationships between patches with a minimum of inductive bias. This makes ViT a model with greater capacity and friendly for a dataset of massive samples like JFT~\cite{jft}.
Since ViT, attention-based methods have deeply reformed vision understanding, and many works attempted to find better attention operators for computer vision~\cite{swin,t2t,pvt,mvit,pit,deformablevit,maxvit}. 
Among these, Swin Transformer~\cite{swin} introduces shifting windows for vision transformers and truncates the attention range inside these windows for better efficiency.
Reducing the number of tokens in the attention operator in the Pyramid or Multi-Scale Vision Transformers (PVT or MViT)~\cite{pvt,mvit} also maintains a better trade-off between performance and computation.

However, despite their ability to stress semantic visual areas dynamically, attention in vision transformers still requires dense per-unit modeling and also introduces computation up to the quadratic level with respect to input resolution.
Zhang {\em et al.}~\cite{Zhang_2021_ICCV} use the sparse attention variant first proposed in NLP~\cite{longformer,reformer} to speed up vision transformers for larger inputs.
Another attempts for faster vision transformer inference is to keep only discriminative tokens and reduce the number of tokens at the inference stage via various strategies and patch score criteria~\cite{evit,ats,tome,rao2021dynamicvit,avit}.

In contrast to sparsifying attention operators or reducing the number of visual tokens in pre-trained models at the inference stage, SparseFormer learns a highly limited number of latent tokens to represent an image and performs transformers solely on these tokens in the latent space efficiently.
Since the number of latent tokens is highly constrained and irrelevant to the input resolution, the computational cost of SparseFormer is low, controllable, and practical.

\noindent\textbf{Perceiver architectures and detection transformers.}
Perceiver architectures~\cite{perceiver,perceiverio} aim to unify different modal inputs using latent transformers.
For image inputs, one or more cross attentions are deployed to transform image grid-form features into latent tokens, and then several Transformer stages are applied to these latent tokens.
Our proposed SparseFormer is greatly inspired by this paradigm, where sparse feature sampling can be regarded as ``cross attention'' between the image space and the latent space.
However, SparseFormer differs from Perceiver architectures in several aspects:
First, in SparseFormer, the ``cross attention'' or sparse feature sampling is performed sparsely and based on direct bilinear interpolation.
SparseFormer does not need to traverse every pixel in an image to build a latent token, while Perceiver architectures require an exhausting traversal.
Second, the number of latent tokens in Perceiver is large (\ie, 512). In contrast, the number of tokens in SparseFormer is much smaller, down to $49$, only $0.1\times$ times compared to Perceiver.
As a result, the proposed SparseFormer enjoys better efficiency than Perceiver.

Recently proposed detection transformers (DETR models)~\cite{detr,deformableDETR,sparsercnn} also use cross attention to directly represent object detections by latent tokens from encoded image features.
SparseFormer shares similar ideas with DETR to represent an image by latent tokens but in a more efficient manner.
SparseFormer does not require a heavy CNN backbone or a Transformer encoder to encode image features first.
The ``cross attention'' layer in SparseFormer is based on the image bilinear interpolation and is also efficient, inspired by the adaptive feature sampling in AdaMixer~\cite{adamixer} as a detection transformer.
Furthermore, SparseFormer does not require localization signals in object detection, and it can be optimized to spatially focus tokens in an end-to-end manner solely with classification signals.

\noindent\textbf{Glimpse models.}
In the nascent days of neural vision understanding research, glimpse models~\cite{rmva,wheretolook} are proposed to imitate human visual perception by capturing several glimpses over an image for recognition. These glimpse models only involve computation over limited parts of an image, and their computation is fixed. While glimpse models are efficient, they are commonly non-differentiable regarding where to look and necessitate workarounds like the expectation-maximization algorithm~\cite{em} or reinforcement learning to be optimized. Furthermore, the performance of glimpse models has only been experimented with promising results on rather small datasets, like MNIST~\cite{mnist}, and the potential to extend to larger datasets has not yet been proven.

Our presented SparseFormer architecture can be seen as an extension of glimpse models, where multiple glimpses (latent tokens) are deployed in each stage.
Since bilinear interpolation is adopted, the entire pipeline can be optimized in an end-to-end manner.
Furthermore, the proposed method has been shown to be empirically effective on large benchmarks.

\section{SparseFormer}\label{sec:method}
In this section, we describe the SparseFormer architecture in detail:
we first introduce SparseFormer as a method for the image classification task and then we discuss how to extend it to video classification with minimal additional efforts.
As SparseFormer performs most of the operations over tokens in the latent space, we begin with the definition of latent tokens.
Then we discuss how to build latent tokens and how to perform recognition based on them.
\FigureArch
\noindent\textbf{Latent tokens.}
Different from dense models, which involve per-pixel or per-patch modeling in the original image space, SparseFormer recognizes an image in a \emph{sparse} way by learning a limited number of tokens in the latent space and applying transformers to them.
Similar to tokens or queries in the Transformer decoder~\cite{DBLP:conf/nips/VaswaniSPUJGKP17}, a token in SparseFormer is an embedding $\mathbf{t}\in \mathbb{R}^{d}$ in the latent space.
To explicitly model the spatial focusing area, we associate each latent token $\mathbf{t}$ in SparseFormer with an RoI descriptor $\mathbf{b}=(x, y, w, h)$, where $x$, $y$, $w$, and $h$ are the center coordinates, width, and height, respectively.
We choose to normalize the components of $\mathbf{b}$ by the image size, resulting in a range of $[0, 1]$.
Thus, a latent token consists of an embedding $\mathbf{t}$ and an RoI descriptor $\mathbf{b}$ as its geometric property.
The entire set of latent tokens in SparseFormer can be described as follows:
\begin{align}
\mathbf{T}=\{(\mathbf{t}_1, \mathbf{b}_1), (\mathbf{t}_2, \mathbf{b}_2), \cdots, (\mathbf{t}_N, \mathbf{b}_N)\},
\end{align}
where $N$ is the number of latent tokens, and both the embedding $\mathbf{t}$ and the RoI $\mathbf{b}$ can be refined throughout the network. The initial $\mathbf{t}$ and $\mathbf{b}$ are learnable parameters of SparseFormer, and their initialization will be described in the experiment section.

\subsection{Building Latent Tokens}
SparseFormer includes two successive parts for building tokens in the latent space, the focusing Transformer and the cortex Transformer.
The focusing Transformer addresses the challenge of sparsely extracting image features, decoding them into latent tokens, and adjusting token RoIs.
The subsequent cortex Transformer accepts these latent token embeddings as inputs and models them with a standard Transformer encoder.

\noindent\textbf{Sparse feature sampling.}
In sampling stages, a latent token in SparseFormer generates $P$ sampling points in the image space according to its geometric property, \ie, the token RoI. These sampling points are explicitly described as image coordinates (\ie, $x$ and $y$) in an image and can be directly used for feature sampling via bilinear interpolation.
Since bilinear interpolation takes $\mathcal{O}(1)$ time for every sampling point, we term this procedure as \emph{sparse feature sampling}.
In contrast, cross attention requires $\mathcal{O}(n)$ time to traverse the input, where $n$ is the input size.
To produce such sampling points, SparseFormer first produces relative offsets for each latent token RoI and then derives absolute sampling locations based on these RoIs.
SparseFormer uses a learnable linear layer to generate a set of sampling offsets for a token conditionally on its embedding $\mathbf{t}$: 
\begin{align}
    \left\{(\bigtriangleup x_i, \bigtriangleup y_i)\right\}_{P} = \mathrm{Linear}(\mathbf{t}),
\end{align}
where the $i$-th sampling offset $(\bigtriangleup x_i, \bigtriangleup y_i)$ indicates the relative position of a sampling point $i$ with respect to a token RoI.
The linear layer contains layer normalization~\cite{ln} over $\mathbf{t}$, which is omitted here and below for more clarity.
These offsets are then translated to absolute sampling locations $(\tilde{x}, \tilde{y})$ in an image with the RoI $\mathbf{b}=(x, y, w, h)$:
\begin{align}
    \begin{cases}
        \tilde{x}_i = x + 0.5\cdot \bigtriangleup x_i\cdot w, \\
        \tilde{y}_i = y + 0.5\cdot \bigtriangleup y_i\cdot h, 
    \end{cases}
\end{align}
for every $i$.
To stabilize training, we perform standard normalization over $\left\{(\bigtriangleup x_i, \bigtriangleup y_i)\right\}_{P}$ by three standard deviations to keep most of the sampling points inside the RoI.

SparseFormer can directly and efficiently extract image features through bilinear interpolation based on these explicit sampling points, without the need for dense traversal over the grid.
Given an input image feature $\mathbf{I}\in\mathbb{R}^{C\times H\times W}$ with $C$ channels, the shape of the sampled feature matrix $\mathbf{x}$ for a token is $\mathbb{R}^{P\times C}$.
The computational complexity of this sparse feature sampling procedure is $\mathcal{O}(N\cdot P\cdot C)$, with the number of latent tokens $N$ given and independent of the input image size $H\times W$.

\noindent\textbf{Adaptive feature decoding.}
Once image features $\mathbf{x}$ have been sampled for a token, the key question for our sparse architecture is how to effectively decode them and build a token in the latent space.
While a linear layer of $\mathbb{R}^{P\times C}\rightarrow\mathbb{R}^d$ can be a simple method to embed features into the latent token, we have found it to be rather ineffective.
Inspired by~\cite{adamixer}, we use the adaptive mixing layer to decode sampled features to leverage spatial and channel semantics in an \emph{adaptive} way.
Specifically, we use a lightweight network $\mathcal{F}:\mathbb{R}^{d}\rightarrow\mathbb{R}^{C\times C+P\times P}$ conditionally on the token embedding $\mathbf{t}$ to produce \emph{adaptive} channel decoding weight $M_c\in \mathbb{R}^{C\times C}$ and spatial decoding weight $M_s\in \mathbb{R}^{P\times P}$. We then decode sampled features $\mathbf{x}$ in the channel axis and spatial axis with adaptive weights in the following order:
\begin{align}
    \left[M_c|M_s\right] &= \mathcal{F}(\mathbf{t}), M_c\in \mathbb{R}^{C\times C}, M_s\in \mathbb{R}^{P\times P}, \\
    \mathbf{x}^{(1)} &= \mathrm{GELU}(\mathbf{x}^{(0)}M_c) \in \mathbb{R}^{P\times C},\\
    \mathbf{x}^{(2)} &= \mathrm{GELU}(M_s\mathbf{x}^{(1)}) \in \mathbb{R}^{P\times C}.
\end{align}
Here, $\mathbf{x}^{(0)}$ represents the sampled feature $\mathbf{x}$, while $\mathbf{x}^{(2)}$ is the output of the adaptive feature decoding process. GELU refers to the Gaussian Error Linear Unit activation function~\cite{gelu}, and learnable biases before GELU are omitted for clarity.
We choose two successive linear layers without activation functions as our $\mathcal{F}$, where the hidden dimension is $d/4$, for efficiency. The final output, $\mathbf{x}^{(2)}$, is passed through a learnable linear layer to dimension $d$ and then added back to the token embedding to update it.
Adaptive feature decoding can be viewed as a token-wise spatial-channel factorization of dynamic convolution~\cite{DBLP:conf/nips/JiaBTG16}, adding convolutional inductive bias for SparseFormer.

The adaptive feature decoding method enables SparseFormer to reason conditionally on the embedding $\mathbf{t}$ about \emph{what a token expects to see}.
Since SparseFormer performs feature sampling and decoding several times, the token embedding $\mathbf{t}$ contains information about \emph{what a token has seen before}.
Thus, a token can infer \emph{where to look} and focus on discriminating foregrounds with the RoI adjusting mechanism, which is described below.

\noindent\textbf{Adjusting RoI.}
A first, quick glance at an image with human eyes is usually insufficient to fully understand its contents.
Our eyes need several adjustments to bring the foreground into focus before we can recognize objects.
This is also the case for SparseFormer.
The RoI of a latent token in SparseFormer can be iteratively adjusted together with the update of its corresponding token embedding.
In a stage, a token RoI $\mathbf{b}=(x, y, w, h)$ is adjusted to $(x', y', w', h')$ in the following way:
\begin{align}
    &x'=x+t_x\cdot w,\hspace{1.3em} y'=y+t_y\cdot h, \\
    &w'=w\cdot\exp(t_w),\hspace{0.5em} h' = h\cdot\exp(t_h),
\end{align}
where $(t_x, t_y, t_w, t_h)$ are normalized adjustment deltas for an RoI, a parameterization adopted from object detectors~\cite{DBLP:conf/nips/RenHGS15}.
The adjustment deltas are produced by a linear layer, which takes the token embedding as input in a token-wise manner:
\begin{align}
    \left\{t_x, t_y, t_w, t_h\right\}=\mathrm{Linear}(\mathbf{t}).
\end{align} 

With the RoI adjusting and sufficient training, SparseFormer can focus on foregrounds after several stages.
It is worth noting that unlike object detectors, we do not supervise SparseFormer with any localizing signals.
The RoI adjustment optimization is achieved end-to-end by back-propagating gradients from sampling points in bilinear interpolation.
Though the bilinear interpolation might incur noisy gradients due to local and limited sampling points, the optimization direction for RoI adjusting is still non-trivial by ensembling gradients of these points according to experiments.

\noindent\textbf{Focusing Transformer.}
In practice, we perform token RoI adjustment first, generate sparse sampling points, and then use bilinear interpolation to obtain sampled features. We then apply adaptive decoding to these sampled features and add the decoded output back to the token embedding vector.
The RoI adjusting, sparse feature sampling, and decoding in all can be regarded as a ``cross attention'' alternative in an \emph{adaptive} and \emph{sparse} way. Together with self-attention over latent tokens and feed-forward network (FFN), we can stack these sampling and decoding as a repeating Transformer stage in the latent space to perform iterative refinement for latent token embedding $\{\mathbf{t}_i\}_N$ and RoI $\{\mathbf{b}_i\}_N$.
We name this repeating Transformer as \emph{the focusing Transformer}.
The focusing Transformer in our design is lightweight, and parameters are shared between repeating stages for efficiently focusing on foregrounds.
These latent tokens are then fed into a standard large Transformer encoder, termed as \emph{the cortex Transformer}.

\noindent\textbf{Cortex Transformer.}
The cortex Transformer follows a standard Transformer encoder architecture except for the first stage with the ``cross attention'' in the focusing Transformer.
As the name implies, the Cortex Transformer behaves like the cerebral cortex in the brain, processing visual signals from the focusing Transformer in a manner more similar to the way the cortex processes signals from the eyes.
Parameters of different cortex Transformer stages are independent.

\subsection{Overall SparseFormer Architecture}
The overall architecture in SparseFormer is depicted in Figure~\ref{fig:arch}.
The input image features are the same and shared across all sampling stages.
The final classification is done by averaging embeddings $\{\mathbf{t}_i\}_N$ over latent tokens and applying a linear classifier following.

\noindent\textbf{Early convolution.}
As discussed, gradients regarding sampling points through bilinear interpolation might be very noisy.
This case is even worse for raw RGB inputs since the nearest four RGB values on the grid are usually too noisy to estimate local gradients.
Thus, it is necessary to introduce early convolution for more interpolable feature maps to achieve better training stability and performance, like ones for vision transformers~\cite{earlyconv}.
In the experiment, the early convolution is designed to be as lightweight as possible.

\noindent\textbf{Sparsity of the architecture.}
It is worth noting that the number of latent tokens in SparseFormer is limited and independent of the input resolution.
The sparse feature sampling procedure extracts features from image feature maps in a non-traversing manner.
Therefore, the computational complexity and memory footprints of transformers in the latent space are \emph{independent} of the input size. So it is reasonable to call our method a sparse visual architecture.
Moreover, the SparseFormer latent space capacity also demonstrates sparsity.
The largest capacity of the latent space, $N \cdot d_c$, is $81\cdot 768$ according to Table~\ref{table:modelConfig}, still smaller than the input image size $3\cdot 224^2$.
This distinguishes SparseFormer from Perceivers~\cite{perceiver,perceiverio}, whose latent capacity is typically $512\cdot 1024$, exceeding the input image size.

It should be also noted that our presented method is quite different from post-training token sparsifying techniques~\cite{evit,ats,tome,rao2021dynamicvit,avit}: SparseFormer, as a new visual architecture, learns to represent an image with sparse tokens from scratch.
In contrast, token sparsifying techniques are usually applied to pre-trained vision transformers.
In fact, token sparsification can also be further applied to latent tokens in SparseFormer, but this is beyond the scope.

\noindent\textbf{Extension to video classification.}
Video classification is similar to image classification but requires more intensive computation due to multiple frames.
Fortunately, SparseFormer architecture can also be extended to video classification with minor additional efforts. Given the video feature $\mathbf{V}\in\mathbb{R}^{C \times T \times H\times W}$, the only problem is to deal with an extra temporal axis compared to $\mathbf{I}\in\mathbb{R}^{C\times H\times W}$.
We associate the token RoI with an extension $(t, l)$, where $t$ is the center temporal coordinate and $l$ is the temporal length of the tube, to make the RoI a tube.
In the sparse feature sampling procedure, an extra linear layer produces temporal offsets, and we transform them to 3D sampling points $\{(\tilde{x}_i, \tilde{y}_i, \tilde{z}_i)\}_P$. Bilinear interpolation is replaced by trilinear interpolation for 4D input data. Alike, the RoI adjusting is extended to the temporal dimension. Other operators like early convolution, adaptive feature decoding, self attention, and FFN remain untouched. For larger capacity for videos, we inflate tokens in the temporal axis by $n_t$ times and initialize their $(t, l)$ to cover all frames, where $n_t$ is much smaller than the frame count $T$ (\eg, $8$ versus $32$).

\section{Experiments}
We conduct experiments on the canonical ImageNet classification~\cite{DBLP:conf/cvpr/DengDSLL009} and Kinetics~\cite{kinetics} video classification benchmarks to investigate our proposed SparseFormer.
We also report our preliminary trials of SparseFormer on downstream tasks, semantic segmentation and object detection, in the supplementary.

\ModelConfig

\noindent\textbf{Model configurations.}
We use ResNet-like early convolutional layers (a $7\times 7$ stride-$2$ convolution, a ReLU, and a $3\times 3$ stride-$2$ max pooling) to extract initial $96$-d image features.
We design several variants of SparseFormer with the computational cost from 2G to $\sim$8G FLOPs, as shown in Table~\ref{table:modelConfig}. We mainly scale up the number of latent tokens $N$, the dimension of the focusing and cortex Transformer $d_f$ and $d_c$, and layers of the cortex Transformer $L_c$.
For all variants, we keep the repeats of the focusing Transformer, $L_f$, to $4$.
We bridge the focusing Transformer and the cortex Transformer with a linear layer to increase the token dimension.
Although the token number is scaled up, it is still smaller than that of conventional vision transformers.
The center of the latent token RoIs and their sampling points are initialized to a grid. The width and height of these RoIs are initialized to half the image.
For the sake of unity in building blocks, the first cortex Transformer stage also performs RoI adjusting, feature sampling, and decoding.
We do not inject any positional information into latent tokens. For more details, please refer to the supplementary.

\ImageNetOneKClassification
\noindent\textbf{Training recipes.}
For ImageNet-1K classification~\cite{DBLP:conf/cvpr/DengDSLL009}, we train the proposed Transformer according to the recipe in~\cite{swin}, which includes the training budget of 300 epochs, the AdamW optimizer~\cite{adamw} with an initial learning rate 0.001, the weight decay 0.05 and sorts of augmentation and regularization strategies.
The input resolution is fixed to $224^2$.
We add EMA~\cite{ema} to stabilize the training.
The stochastic depth (\ie, drop path)~\cite{droppath} rate is set to 0.2, 0.3, and 0.4 for SparseFormer-T, -S, and -B.

To pre-train on ImageNet-21K~\cite{DBLP:conf/cvpr/DengDSLL009}, we use a subset suggested by~\cite{ridnik2021imagenet} due to the availability of the \texttt{winter 2021 release}. We follow the training recipe in~\cite{swin} and use a similar recipe to 1K classification but with 60 epochs, an initial learning rate $2\times 10^{-3}$, weight decay 0.05, and drop path 0.1.
After pre-training, we fine-tune models with a recipe of 30 epochs, an initial learning rate $2\times 10^{-4}$ with cosine decay and weight decay $10^{-8}$. 

We adopt the ImageNet pre-trained models to initialize parameters for training on Kinetics-400. Since our architecture is endurable to large input sizes, the number of input frames is set to $T=32$. Formally, 32 frames are sampled from the 128 consecutive frames with a stride of 4. We mildly inflate initial latent tokens by $n_t=8$ times in the temporal axis to cover all input frames.
We use AdamW~\cite{adamw} for optimization with 32 GPUs, following the training protocol in~\cite{mvit}. We train the model for 50 epochs with 5 linear warm-up epochs. The mini-batch size is 8 video samples per GPU.
The learning rate is set to $5\times 10^{-4}$, and we adopt a cosine learning rate schedule~\cite{coslr}. For evaluation, we apply a 12-view testing scheme (three 3 spatial crops and 4 temporal clips) as previous work~\cite{liu2021video}.
\Ablation

\subsection{Main Results}
\noindent\textbf{ImageNet-1K classification.}
We first benchmark SparseFormer on the ImageNet-1K classification and compare them to other well-established methods in Table~\ref{table:imagenetclassification}. 
SparseFormer-T achieves 81.0 top-1 accuracy on par with the well-curated dense transformer Swin-T~\cite{swin}, with less than half FLOPs of it (2.0G versus 4.5G) and 74\% higher throughput (1270 versus 726).
The small and base variants of SparseFormer, SparseFormer-S, and -B also maintain a good balance between the performance and actual throughput over highly-optimized CNNs or transformers.
We can see that Perceiver architectures~\cite{perceiver,perceiverio}, which also adopt the latent transformer as ours, incur extremely large FLOPs and have impractical inference speed due to a large number of tokens (\ie, $512$) and dense cross-attention traversal.

\ScalingUp
\noindent\textbf{Scaling up SparseFormer.}
We scale up the base SparseFormer variant in Table~\ref{table:twentyonekandscalingup}.
We first adopt ImageNet-21K pre-training, and it brings 1.0 top-1 accuracy improvement.
Then we investigate SparseFormer with large input resolution fine-tuning.
Large resolution inputs benefit SparseFormer (0.5$\uparrow$ for $384^2$) only extra 5\% FLOPs.
Moreover, we try a more aggressive way to scale up the model by {reinitializing} tokens (\ie, embeddings and RoIs) with a more number \emph{in fine-tuning}, and find it with better results.
We leave further scaling up to future work.

\noindent\textbf{Kinetics-400 classification.}
\VideoUnderstanding
We also investigate the extension of SparseFormer to the video classification task. Results on the Kinetics-400 dataset are reported in Table~\ref{table:video}.
Our VideoSparseFormer-T achieves the performance of well-established video CNNs (I3D or SlowFast) with a much lower computational burden.
Surprisingly, our VideoSparseFormer-S pre-trained on ImageNet-1K even surpasses the Transformer-based architectures pre-trained on ImageNet-21K, like TimeSFormer~\cite{timesformer} and ViViT~\cite{vivit}. Furthermore, our VideoSparseFormer-S pre-trained on ImageNet-21K can improve the performance to 79.8 with only 74 GFLOPs.

\subsection{Ablation Study}
In this section, we ablate key designs in the proposed SparseFormer.
Limited by computational resources, we resort to SparseFormer-T on ImageNet-1K classification.

\noindent\textbf{The number of latent tokens.}
The number of latent tokens $N$ is a key hyper-parameter of SparseFormer as it controls the latent space capacity.
Table~\ref{tab:tokennumber} shows the performance with different numbers of latent tokens. 
The performance of SparseFormer is highly determined by the number of tokens $N$, and so is the computational cost.
We can see when increasing $N$ to 81, SparseFormer-T reaches the performance of SparseFormer-S.
As there are no dense units in the latent space, the number of tokens serves a crucial role in the information flow in the network.
We are still in favor of fewer tokens in the design of SparseFormer for efficiency.

\noindent\textbf{Focusing stages.}
Given a limited number of latent tokens, SparseFormer attempt to focus them on foregrounds by adjusting their RoI and sample corresponding features to make a visual recognition.
Table~\ref{tab:focusingstage} investigates repeats of the parameter-shared focusing Transformer.
The `nil' stands for fixing token RoIs to specific areas and making them unlearnable. Sampling points of tokens in the `nil' row are also set to fixed spatial positions.
The adjusting RoI mechanism and its repeats are vital to the SparseFormer performance.

\noindent\textbf{Sparsity of sampling points.}
The other important factor for the sparsity of the proposed method is the number of sampling points in the feature sampling. Table~\ref{tab:samplingpoints} shows the ablation on this.
Compared to increasing the number of latent tokens (\eg, $49\to 64$, 30\% up, 81.4 accuracy), more sampling points are not economical for better performance, and 81.3 accuracy needs 77\% more ($36\to 64$) sampling points.
To maintain a tradeoff between sparsity and performance, we choose 36 sampling points as our default.

\noindent\textbf{Image features and how to decode sampled features.}
Table~\ref{tab:imagefeatures} investigates input image features to be sampled into the latent space in SparseFormer.
As discussed before, input image features can be in the raw RGB format, but we find it hard to train\footnote{Attempts of a preliminary design of SparseFormer with the raw RGB input are with about 60 top-1 accuracy.}.
We also ablate it with ViT-like embedding layer~\cite{vit}, namely, patchifying but keeping the spatial structure, and find it worse than the ResNet-like early convolutional image feature.
Table~\ref{tab:howtodecode} ablates how to decode sampled features for a token. The static mixing uses the static weights, which are not conditional on token embedding, to perform mixing on sampled features.
We can find that the adaptive mixing decoding design is better.  

\noindent\textbf{Inflation of latent tokens on video classification.}
We also investigate the inflation rate of tokens on videos. Intuitively, video data with multiple frames need more latent tokens than a static image to model.
Results in Table~\ref{table:videoInflation} show this.
Note that the input video has 32 frames, but the token inflation rate 8 is already sufficient for the favorable performance of VideoSparseFormer.
As a contrast, dense CNNs or Transformers usually require at least exactly $\#\textrm{frames}$ times the computational cost if no temporal reduction is adopted.
This also validates the sparsity of the proposed SparseFormer method on videos.
\VideoUnderstandingAblationOnInflation

\subsection{Visualizations}
\FigureVis
As discussed in Section~\ref{sec:method}, we argue that SparseFormer, with the RoI adjusting and sparse feature sampling, can focus on foregrounds by reasoning about where to look.
To show this, we perform visualizations of token sampling points across different sampling stages in Figure~\ref{fig:vis}.
We apply kernel density estimation (KDE)~\cite{kde} spatially about sampling points with top-hat kernels to obtain the sampling density map.
We can find that SparseFormer initially looks at the image in a relatively uniform way and gradually focuses on discriminative details of foregrounds. It is worth noting that SparseFormer is supervised with only classification signals, and it can roughly learn \emph{where discriminative foregrounds are} by \emph{weak supervision}.

\section{Conclusion}
In this paper, we have presented a neural architecture, SparseFormer, to perform visual recognition with a limited number of tokens along with the Transformer in the latent space.
To imitate human eye behavior, we design SparseFormer to focus these sparse latent tokens on discriminative foregrounds and make a recognition sparsely.
As a very initial step to the sparse visual architecture, SparseFormer consistently yields promising results on challenging image classification and video classification benchmarks with a good performance-throughput tradeoff.
We hope our work can provide an alternative way and inspire further research about sparse visual understanding.

\clearpage
\begin{figure*}[h!]
\begin{center}
    \centering
    \captionsetup{type=figure}
    \includegraphics[width=.8\textwidth]{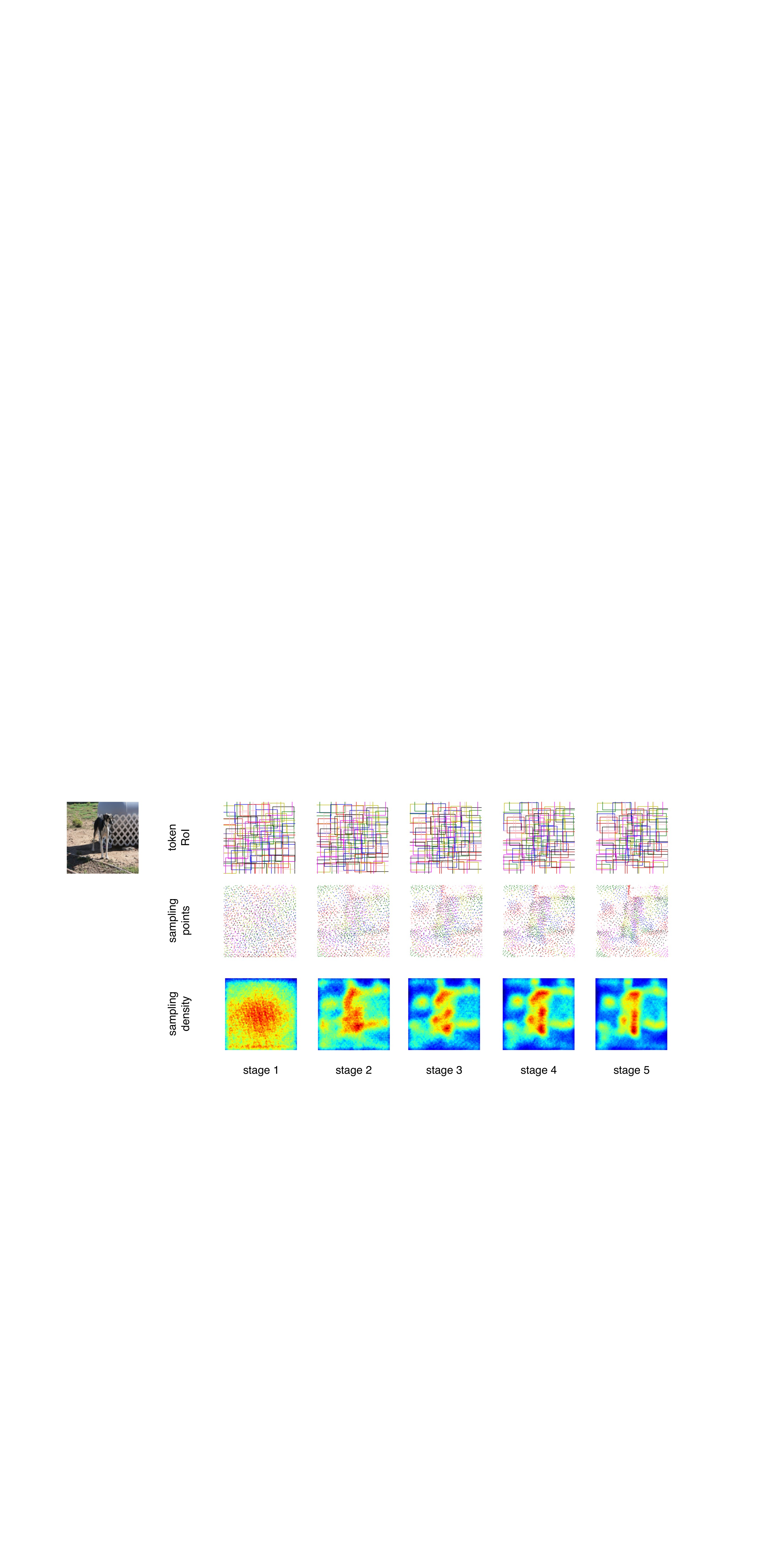}
    \captionof{figure}{\textbf{More visualizations} of token RoIs, their sampling points, and density across sampling stages in SparseFormer-S ($64$ tokens). RoIs and sampling points of different tokens are colored with different colors. Better view with zoom-in.}\label{fig:one}
    
\end{center}
\end{figure*}
\section*{Appendix}

\subsection*{A.1. SparseFormer itself as Object Detector}
Since the SparseFormer architecture produces embedding and RoI together for a token given an image, it is natural to ask whether SparseFormer \emph{per se} can perform object detection task?
The answer is \emph{yes}.
In other words, we can train a SparseFormer model to detect objects \emph{without making any architectural changes} by simply adding a final classifier and a RoI refining layer upon it.

Specifically, we follow the training strategy of DETR~\cite{detr} to train a SparseFormer-S for object detection.
We adopt the ImageNet-1K pre-trained SparseFormer-S in the main paper.
We first inflate the number of latent tokens to $400$ by re-initializing token embeddings to the normalization distribution and the center of token RoIs to the uniform distribution on $[0, 1]$.
The RoI height and width is $0.5\times 0.5$ still.
We use a fixed set of $100$ latent tokens to detect objects. The other tokens, which are not used for detection, aim to enrich the semantics in the latent space.
We do not change the matching criterion together with the loss function of DETR and we train for $300$ epochs.
The final classifier is simply one-layer FC layer and the RoI refining layer is 2-layer FC layer, also following DETR.
The final refining of RoIs is performed in the same way as RoI adjustment in the main paper. The result is shown in Table~\ref{tab:det}.
\Det

Although the performance of SparseFormer is currently inferior to DETR, it is important to note that this is very preliminary result and we do not add any additional attention encoders or decoders to SparseFormer for object detection.
Actually, SparseFormer can be considered a decoder-only architecture for object detection if we treat the early convolution part as the embedding layer.

\subsection*{A.2. SparseFormer Performing Per-Pixel Task}
SparseFormer learns to represent an image by limited tokens in the latent space and outputs token embeddings with their corresponding RoIs. It is appealing to investigate whether SparseFormer can perform per-pixel tasks, like semantic segmentation. Yet, SparseFormer itself cannot perform dense tasks since it outputs discrete token set.
However, we can \emph{restore} a dense structured feature map from these discrete tokens by the vanilla cross-attention operator and build a final classifier upon the dense feature map.

Specifically, to perform semantic segmentation, we use a location-aware cross-attention operator to map latent token embeddings back to the structured feature map, whose height and width is one fourth the input image (namely, stride $4$, $H/4$ and $W/4$).
The location-aware cross-attention is the vanilla cross-attention with geometric prior as biases in attention matrix:
\begin{align*}
    {\rm Attn}(Q_{ds}, K_{lt}, V_{lt})={\rm Softmax}(Q_{ds}K_{lt}^T/\sqrt{d}+B)V_{lt},
\end{align*} 
where $Q_{ds}\in \mathbb{R}^{N_{ds}\times d}$ is the query matrix for the dense map ($N_{ds}=H/4*W/4$), $K_{lt}, V_{lt}\in \mathbb{R}^{N\times d}$ is the key and value matrix for the latent tokens, 
\begin{align*}
    B_{i,j}=-(\frac{x_{ds, i}-x_{lt, j}}{w_{lt, j}})^2-(\frac{y_{ds, i}-y_{lt, j}}{h_{lt, j}})^2
\end{align*},
where $(x_{lt}, y_{lt}, w_{lt}, h_{lt})$ is the RoI descriptor for a latent RoI and $(x_{ds}, y_{ds})$ are $x$ and $y$ coordinates for a unit on the dense feature map.
In our current design, the input dense map to cross attend latent tokens is the early convolved feature map, which has the same height and width $H/4$ and $W/4$.
We put two $3\times 3$ convolution layers and a following classifier on the restored feature map following common pratice. The results are shown in Table~\ref{tab:seg}.

\Seg

We also inflate the number of latent tokens in the semantic segmentation as we do in object detector for better performance.
The performance of SparseFormer-T with 400 latent tokens is near the well-established Swin~\cite{swin} and UperNet~\cite{UperNet} but with merely $1/8$ Swin-T's GFLOPs.
This validates that our proposed SparseFormer can perform per-pixel task and is suitable to handle high resolution input data with limited latent tokens.

\subsection*{A.3. Model Initializations in Details}
We initialize all weights of linear layers in SparseFormer unless otherwise specified below to follow a truncated normalization distribution with a mean of $0$, a standard deviation of $0.02$, and a truncated threshold of $2$.
Biases of these linear layers are initialized to zeros if exisiting.

Sampling points for every token are initialized in a grid-like shape ($6\times 6$ for $36$ sampling points by default) by zeroing weights of the linear layer to generate offsets and setting its bias using \texttt{meshgrid}.
Alike, we initialize the center of initial token RoIs (as parameters of the model) to the grid-like (\eg, $7\times 7$ for $49$ SF-Tiny variant) shape in the same way.
The token's height and width are set to half of the image's height and width, which is expressed as $0.5 \times 0.5$.
We also try other initializations for tokens' height and width in Table~\ref{tab:box_init}.
For training stability, we also initialize adaptive decoding in SparseFormer following~\cite{adamixer} with an initial Xavier decoding weight~\cite{xavier}.
This initialization makes the adaptive decoding behaves like unconditional convolution (weights not dependent on token embeddings) at the beginning of the training procedure.

\BoxInit

For alternative ways for token height and width initializations, we can find that there is no significant difference between the `half' and `cell' initializations.
We prefer the `half' initialization as tokens can see more initially.
However, setting all token RoIs to the whole image, the `whole' initialization, is lagging before other initializations.
We suspect that the model is unable to differentiate between different tokens and is causing training instability due to identical RoIs and sampling points for all tokens.

\subsection*{A.4. Visualizations}
\textbf{More visualizations on RoIs and sampling points.}
In order to confirm the general ability of SparseFormer to focus on foregrounds, we present additional visualizations in Figure~\ref{fig:one} and \ref{fig:two} with ImageNet-1K~\cite{DBLP:conf/cvpr/DengDSLL009} validation set inputs.
Note that these samples are not cherry-picked.
we observe that SparseFormer progressively directs its attention towards the foreground, beginning from the roughly high contrasting areas and eventually towards more discriminative areas.
The focal areas of SparseFormer adapt to variations in the image and mainly concentrate on discriminative foregrounds when the input changes.
This also validate the semantic adaptability of SparseFormer to different images.

\textbf{Visualizations on specific latent tokens.}
We also provide visualizations of specific latent tokens across stages to take a closer look at how the token RoI behaves at the token level.
We choose 5 tokens per image that respond with the highest values to the ground truth category. To achieve this, we remove token embedding average pooling and place the classifier layer on individual tokens. The visualizations are shown in Figure~\ref{fig:token}.
We can observe that the token RoIs progressively move towards the foreground and adjust their aspect ratios at a mild pace by stage.

\textbf{Visualizations on disturbed input images.}
We also show visualizations on disturbed input images in Figure~\ref{fig:distu}, where images are either random erased or heavily padding with zero values or reflection.
We can see that although SparseFormer initially views the image in a almost uniform way, it learns to avoid sampling in uninformative areas in subsequent stages.
This illustrates the robustness and adaptability of SparseFormer when dealing with perturbed input images.
\newpage
{\small
\bibliographystyle{ieee_fullname}
\bibliography{egbib}
}

\FigureSuppOne
\FigureSuppToken
\FigureSuppDis

\end{document}